\documentclass[conference]{IEEEtran}
\usepackage[T1]{fontenc}
\usepackage[utf8]{inputenc}
\usepackage{cite}
\usepackage{amsmath,amssymb,amsfonts}
\usepackage{graphicx}
\usepackage{textcomp}
\usepackage{xcolor}
\usepackage{url}
\usepackage{array}
\renewcommand{\thetable}{\arabic{table}}
\renewcommand\IEEEkeywordsname{Keywords}
\usepackage[normalem]{ulem}

\begin{document}

\title{CLEAR: Converting PDFs Locally into Enriched, Agent-Ready Markdown for Reliable Retrieval}

\author{\IEEEauthorblockN{Kostas Ordoumpozanis}
\IEEEauthorblockA{\textit{Department of Cultural Technology and Communication} \\
\textit{University of the Aegean} \\
Myteline, Greece}
\and
\IEEEauthorblockN{George A. Papakostas}
\IEEEauthorblockA{\textit{Applied Artificial Intelligence Laboratory (AAI-Lab)} \\
\textit{Department of Informatics, Democritus University of Thrace} \\
Kavala, Greece}
}

\maketitle

\begin{abstract}
Researchers and companies now base decisions on answers from AI systems that read their documents. Three problems make these answers less reliable than they look. First, a fact may appear partly in the text and partly in a figure or table, or only in the figure or table. If the AI system misses that part, its answer may be wrong with no visible sign. Second, AI answers become new content that future models may learn from, so the missing information may become a new kind of bias. Third, the documents may hold private data that must not leave the owner's computer, so the whole process must run locally. Our goal is to give the AI system the full content of a document, figures included, on one local machine. We build a local pipeline that converts a PDF into Markdown and describes each figure as text inside a tag. We test it on research papers. Five AI models answer 19 two-hop questions whose answers exist only inside figures, and we compare the original PDF with the parsed Markdown as input. The parsed Markdown gives higher accuracy than the PDF in four of six matched pairs, and in one pair accuracy rises from 89.5\% to 100\%. Each answer costs 1.6 to 18.1 times less with the parsed Markdown than with the PDF. In six setups that read the parsed Markdown, three of the five models (Sonnet, Opus and Haiku) could also open the original figure image when the Markdown text did not hold enough. Each of these six setups answered the 19 questions in three runs, so the six setups gave 342 answers in total. In 258 of these answers, the Markdown text alone gave the correct answer, and no image was opened. In 58 answers, the model opened the image, and 56 of these 58 answers were correct. We tested only research papers, but the same local process may also apply to business documents. This work is the first part of a local harness system for research.
\end{abstract}


\section{Introduction}

A research PDF holds text, tables, charts, and diagrams. An AI system that reads this PDF needs the data in each part, not only in the text. Once it reaches the model, the quality of this data determines the quality of its answer. Researchers and companies now base decisions on such answers. A fact may appear partly in the text and partly in a figure or table, or only in the figure or table. If the system misses that part, its answer may be wrong with no visible sign. Document parsing turns a PDF into one clean, structured file, so a model reads accurate data instead of a raw page image. This is an active research problem [1,2]. A parser must find the correct headings, keep the correct reading order, and turn each figure into words a language model can use. This structured output can be the input to a knowledge base, used by retrieval-augmented generation to answer questions from real documents [3]. This research continues a previous work [4]. That work gave Google NotebookLM the same source in different forms. Its answers ignored facts that existed only inside an image. These misses disappeared once the image was transcribed to text before the tool read it. These misses matter beyond one answer, because AI answers also become new content, and future models may learn from it. Models trained on text made by earlier models lose information [5]. Missing information may then become a new kind of bias.

Data quality is not the only reason for this work. A research or business document may also hold private data, and some parsing systems send the document to a large, closed model through a cloud API. A closed cloud model runs outside the user's own machine, so the data must leave the machine to reach it. For private data, this is not acceptable. Running the model locally, on the user's own hardware, keeps the data inside the machine at every step [6]. A local model must therefore match the data quality of a closed one.

Structure is a third reason. Some new memory systems for AI agents store what they learn in a graph, not in free text. Structure makes retrieval cheaper and more exact [7]. Cost is a fourth reason. An agent that needs many rounds to reach a good answer spends tokens on each round [8]. Better input data from the start may lower this cost.

Parsing systems fall into a few categories. A rule-based system uses fixed code, with no model, to find headings and layout. An OCR-plus-layout system reads the page image. It finds the text with optical character recognition. A separate model finds the layout, such as one that fuses text and layout in a single pre-trained model [9]. A vision-language model reads the page image directly and writes structured text for every part. Each category may trade accuracy against local compute cost, so a system for one local GPU must choose with care.

This paper makes six contributions. (1) It builds a fully local parsing pipeline. (2) It creates a quantized version of the Popo [10,11] model. (3) It proposes the <image> XML tag, a common form of image conversion. It gives an agent two levels of depth: description and URL. (4) It shows that a parsed Markdown file, before it is used by an LLM, can raise accuracy and reduce cost. (5) It compares models on both accuracy and cost. (6) The agent holds two levels of information and decides when it needs to go deeper. The paper shows that this ability can improve the accuracy of the agent's results.

Section 2 gives the method steps in order. Section 3 shows the system architecture. Section 4 describes the four databases we built. Section 5 compares the parsers and shows how we fixed the headings. Section 6 explains how the pipeline turns each image into text. Section 7 defines the <image> XML block. Section 8 describes the MCP server and the skill. Section 9 evaluates the parsed Markdown against the PDF. Section 10 reports the time cost of the pipeline. Section 11 lists the limitations, and Section 12 concludes the paper. This pipeline is the first part of a larger, local harness system.

\section{Methodology}

This paper follows one path. Each step solves the problem the step before it left open. First, we gathered thirty diverse papers from five fields, all open access (Section 4). Second, we built a heading ground truth, to grade heading correction later (Section 4). Third, we compared several parsers, to choose one to build on (Section 5). Fourth, we quantized a heading-correction model. It fits one local GPU and solves the heading errors of the previous step (Section 5). Fifth, we wrote a small Python fix for specific headings the pipeline still missed (Section 5). Sixth, we built two more datasets (Section 4). One holds images from the papers, to check classification and description. The other holds questions, to grade a model's own answers. A two-hop question needs two facts from two different images. Seventh, we worked on parsing images to text (Section 6). We tuned the classification prompt. Then we tuned the image-to-text prompt for each type, including tables. Eighth, we designed the <image> XML tag for AI agents (Section 7). It carries each figure's type, link, and text in one place. Ninth, we wrapped the whole pipeline as one local MCP server, with its own tools, resources, and skills (Section 8). This gives AI agents one clear way to parse documents. Tenth, and last, we ran the downstream test through that server, three times per setup, to grade the result (Section 9).

\section{System architecture}

The system turns a PDF into a Markdown file. The pipeline runs five stages in order. Figure 1 shows the flow.

The pipeline converts the PDF into Markdown, using MinerU [12]. This stage extracts the text and the images. It also writes a first, draft version of the headings.

The pipeline corrects errors in the text and repairs the draft headings, using Popo in a quantized version we made and published on Hugging Face.

Inside one outer stage called ``Image to text,'' the pipeline classifies each figure, using Qwen3.5-4B [13]. This step sets the figure's type: \textit{chart}, \textit{diagram}, \textit{photo}, or \textit{table}.

Inside the same outer stage, the pipeline writes a text description of each figure using Qwen3.5-4B. A table that MinerU did not parse the first time goes back to MinerU, this time alone, so MinerU can convert it to a <table> tag. If MinerU returns an empty tag, the pipeline passes the table image to Qwen3.5-4B, with a prompt that converts an image to a <table> tag.

The pipeline writes image tags. It places each figure's description into one <image> block, in place of the image reference.

The final output is one Markdown file. The file holds text only. Every figure appears as an <image> tag. The headings are corrected.

One constraint shapes this whole design. Every stage runs on one local machine, with one GPU and 12 GB of memory. No data leaves the machine. To effectively handle the memory limitation, the system is designed to run one model at a time. The entire system is delivered in Docker that auto creates the needed containers from the images. At the end of each task, it removes the container to free the memory for the next step.

\begin{figure}[!t]
\centering
\includegraphics[width=0.85\columnwidth]{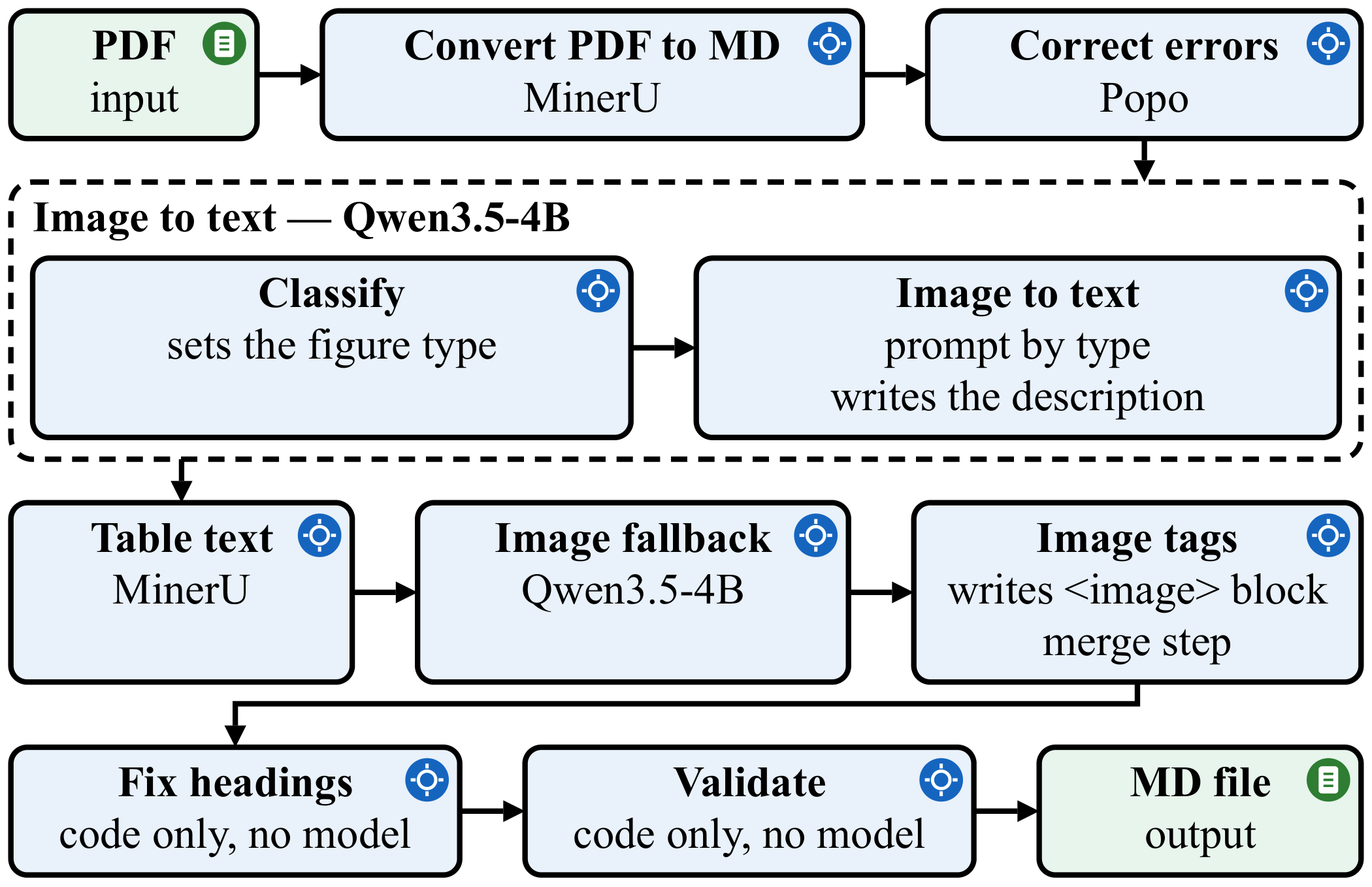}
\caption{The flowchart shows the nine stages of our pipeline in order to convert a PDF to Markdown.}
\label{fig:1}
\end{figure}

\section{Databases}

This paper needs fixed, checked data to test and score every part of the pipeline. This section covers the four databases we built for that.

\subsection{Selecting the Base Research Papers}

The test set is 30 open-access papers, 6 from each of 5 fields: Computer Science, Engineering, Medicine, Physics and Astronomy, and Social Sciences. Five fields give the test set five different paper layouts, so no result in this paper rests on one single layout. A script drew random candidates from OpenAlex [14], kept only papers about their field, written in English, 6 to 30 pages long, with at least 3 figures, downloaded the PDF of each kept paper, and skipped a paper it could not reach. Then the authors checked every kept paper by eye for real text (not a scanned picture) and real English. This process kept 30 papers from 136 candidates.

\subsection{Building the Correct Heading List}

The authors manually extracted the correct heading list for each of the 30 papers, the key this study uses to score every parser. A checker app shows a draft heading list beside the paper's PDF, one paper at a time; the authors read each heading against the page, set its level, and accept or correct it. Section 5 uses this corrected, final list.

\subsection{Labeling the Figure Types}

The authors labeled 100 figures cut from the 30 papers. They gave each one a type and a short description, by eye, so later figure-classification scores have a human-made answer to check against. The final counts are 36 charts, 31 diagrams, 18 photos, 10 other, 3 screenshots, and 2 tables.

\subsection{Building the Figure-Only Questions}

We built test questions that an LLM can answer only with facts from figures. We built a small agent system with two agents per PDF, a creator and a validator, and a human check at the end. The creator proposes a question and explains why it passes. The validator checks whether the question can be answered without the facts from the image. First, the system keeps only papers with 2 or more charts, tables, or diagrams. The creator reads the <image> tag's text description in the paper's converted file and checks it against the rest of the text. If that information also appears in the text, it rejects it and moves on. If not, the agent tries to build a two-hop question: one that needs two facts from two different images to answer. The validator then tests this question with the text alone, leaving the image data out. If the text-only answer fails, the question is a valid figure-only question. If it does not fail, the validator rejects the question. Section 9 uses only this two-hop question type. The creator and validator used Sonnet 5.5 [15] with high effort. At the end, a human cross-validated each question, and we received 19 two-hop questions (each needs at least 2 facts, one per image).

Together these four databases give every later section a fixed, human-checked reference to test and score against.

\section{Parser comparison}

\begin{figure}[!t]
\centering
\includegraphics[width=0.80\columnwidth]{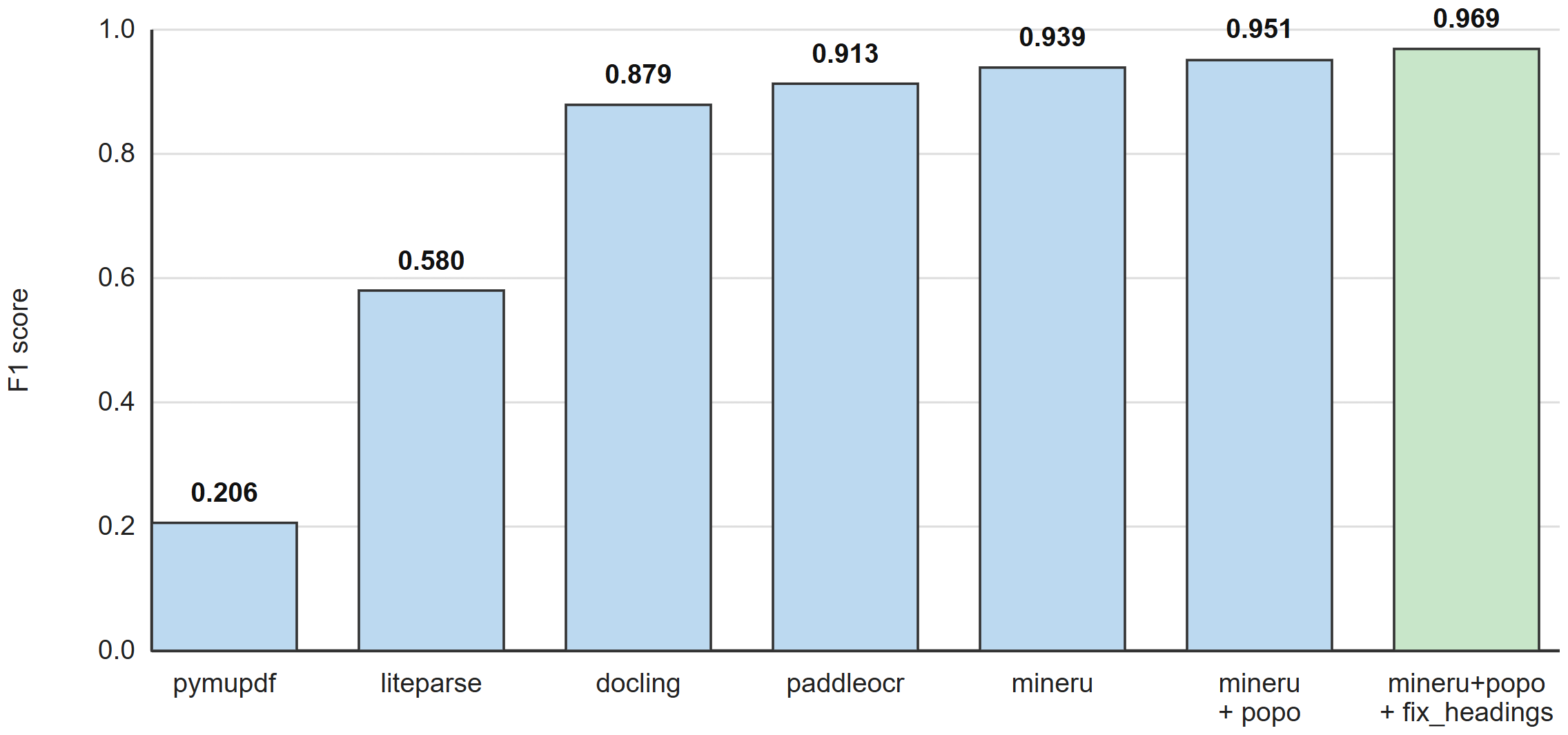}
\caption{The bar chart shows the score of each parser, in the order tested: pymupdf, LiteParse, Docling, PaddleOCR-VL, MinerU, MinerU + Popo, and the final MinerU + Popo + heading fix.}
\label{fig:5}
\end{figure}

We tested many parsers, from plain code to a vision-language model, to choose one parser. Figure \ref{fig:5} shows the score of every step and the final accuracy of heading conversions. We start on 4 of the 30 papers. After we select the method, we run all 30.

Four tools use code only, with no model. pymupdf4llm [16] uses the PDF’s table of contents, or font size when there is none. LiteParse [17] uses no OCR step and no images. Docling [18] uses its default heading hierarchy. PaddleOCR-VL [19] runs OCR with a vision-language model, in a Docker container. MinerU converts the PDF with its own model and writes a draft version of the headings. Its F1 score is 0.939, the highest of the five.

The F1 score measures whether the parser finds the right headings. A second score, level accuracy, checks whether a found heading has the right depth (\#, \#\#, \#\#\#,...). On this score MinerU is weak: level accuracy is only 0.550 on all 30 papers. Without a fix, MinerU gives only two depths (title and section). This led us to Popo, a local language model that corrects the text and repairs MinerU's draft headings. Popo does not change which headings are found: matched headings stay at 513. It does fix their level. With our own quantized Popo build, the level accuracy rises from 0.550 to 0.965.

\begin{figure}[!t]
\centering
\includegraphics[width=0.62\columnwidth]{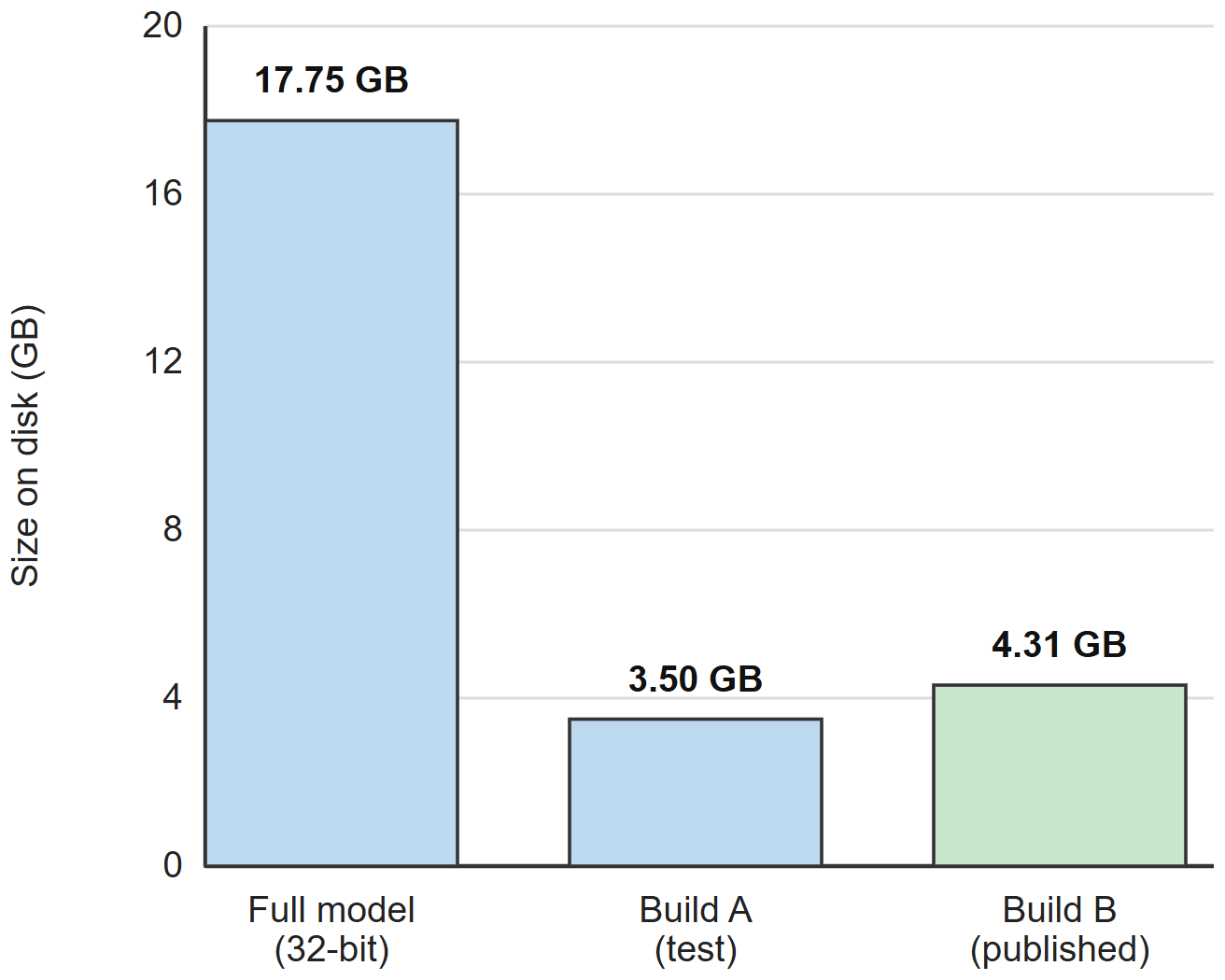}
\caption{The bar chart shows the drop in size on disk from the full Popo model to the published quantized build.}
\label{fig:6}
\end{figure}

\begin{figure}[!t]
\centering
\includegraphics[width=0.70\columnwidth]{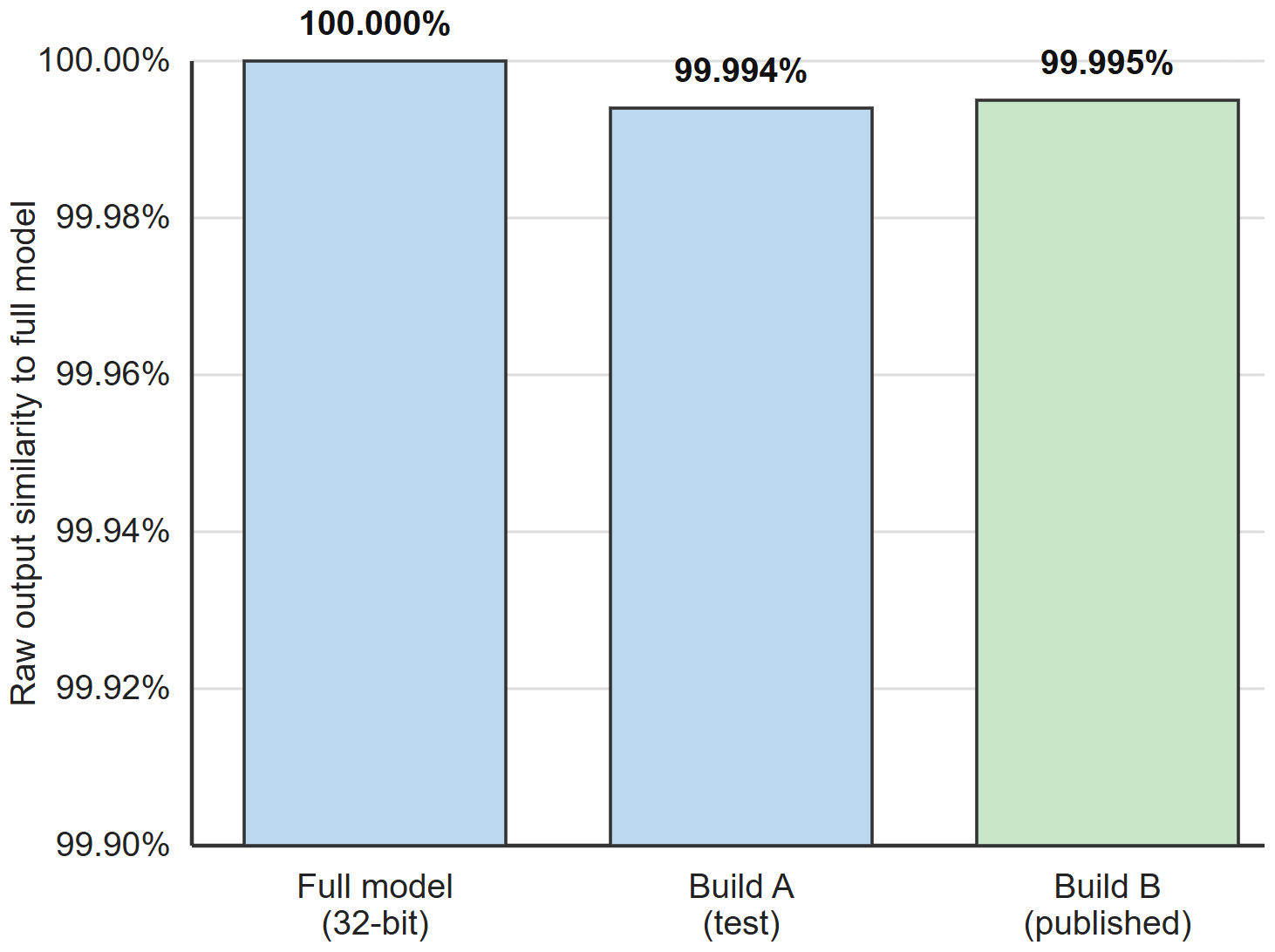}
\caption{The bar chart shows how close the quantized Popo build stays to the full model: raw output similarity of 99.994\% (test build) and 99.995\% (published build), against the full model's 100.000\% match to itself.}
\label{fig:7}
\end{figure}

To run Popo on the local machine, we had to reduce its size. The full model needs 17.75 GB. Our GPU has 12 GB. We tested two quantized builds. Build A (test) kept 2 skip patterns and 105 layers in 16-bit. Build B (published) kept 10 skip patterns and 130 layers in 16-bit. We also measured an error-ranked skip list. We rejected it, because its error stayed flat for every layer. We publish Build B as kordou/MinerU-Popo-NF4 on Hugging Face. Figure~\ref{fig:6} shows that Build B is much smaller. Figure~\ref{fig:7} shows that Build B stays close to the full model. We never scored the full, 32-bit model on all 30 papers. Only Build B, the one the pipeline runs, has an F1 score against the key: 0.951 (Figure~\ref{fig:5}).

Many remaining errors are on the ``Abstract'' and ``Keywords'' lines of the paper. Most parsers may not recognise these lines as a heading. The 30 papers hold 542 headings. MinerU + Popo missed 29 of them. Of these 29, 8 are Abstract, missed in 8 of the 30 papers. 4 are Keywords, missed in 4 of the 30 papers.

We built a small Python script with a regex rule. The script recognises ``Abstract'' and ``Keywords'' at the start of a line. It makes each one a heading (\#\#), if it is not one already. Figure \ref{fig:5} shows the final score: an accuracy of 0.969, counting only a missing heading as an error.

\section{Image Parsing}

This pipeline turns each image in a paper into text. It uses two models: Qwen3.5-4B, a vision-language model, and MinerU, the same parser that reads the page itself. The pipeline sorts each image into one type first: table, chart, diagram, photo, screenshot, or other. Table~\ref{tab:types} lists each type. One general prompt tells Qwen3.5-4B to describe the image as text, numbers included. The prompt holds the image type as a placeholder, so one prompt works for each type. A photo, a screenshot, or an image of another type has no fixed structure, so one general prompt is enough for them. We improved the prompt for charts and diagrams. Tables go to the MinerU parser, with Qwen3.5-4B as a fallback.

\begin{table}[!t]
\caption{Image types, and the prompt used for each type.}
\centering
\footnotesize
\begin{tabular}{p{0.14\columnwidth}p{0.50\columnwidth}p{0.24\columnwidth}}
\hline
\textbf{Type} & \textbf{Meaning} & \textbf{Prompt used} \\
\hline
table & A grid of rows and columns that holds only text or numbers. It has no shapes, colours or pictures. & MinerU engine then Qwen3.5-4B with a tuned prompt \\
chart & A graph with axes that plots data with only lines, bars, points or areas. Examples: bar, line, pie, scatter, area, histogram, box plot, confusion matrix, heatmap. & Prompt improved for charts \\
diagram & A schematic, flowchart, architecture drawing, node-and-arrow graph, map, timeline, mind map, word cloud, or other explanatory illustration. A graph with axes that also holds pictures or icons is a diagram. & Prompt improved for diagrams \\
photo & Only a realistic picture, as taken by a camera. A rendering is not a photo. & General prompt \\
screenshot & A capture of a screen: a user interface, an app, a website, a terminal, code, or a page full of text. & General prompt \\
other & A logo, an icon, or anything that does not clearly fit table, chart, diagram, photo, or screenshot. & General prompt \\
\hline
\end{tabular}
\label{tab:types}
\end{table}

\subsection{The classification prompt}

Each image needs a type before the pipeline can pick its chart or diagram prompt. We changed the classification prompt five times. Each change aimed to raise accuracy and remove a defect of the version before it. Table~\ref{tab:defects} names each defect.

\begin{table}[!t]
\caption{Defects of the classification, chart, and diagram prompts.}
\centering
\footnotesize
\begin{tabular}{p{0.16\columnwidth}p{0.24\columnwidth}p{0.46\columnwidth}}
\hline
\textbf{Prompt} & \textbf{Defect} & \textbf{What it means} \\
\hline
Classification & Wrong table call & The model calls a picture or a shape a table. \\
Classification & Chart/diagram mix-up & The model calls a chart a diagram, or a diagram a chart. \\
Classification & Missed confusion matrix & The model calls a confusion matrix or a heatmap a table, instead of a chart. \\
Classification & Forced match & The model puts a logo or a screenshot into a table, chart, or diagram, because the prompt gives it no other class to choose. \\
Chart & Wrong axis label & The model gives one panel the axis label of a different panel. \\
Chart & Invented meaning & The model explains a symbol's meaning that the chart does not state. \\
Chart & Axis limit as data & The model reports the axis range as if it were a plotted value. \\
Chart & Lost boxplot label & The model does not report the boxplot's own axis label. \\
Chart & Does not stop & The model keeps writing after it has described every chart. \\
Chart & Repeats itself & The model writes the same sentence many times in one answer. \\
Chart & Made-up values & The model writes a number that the chart does not show. \\
Chart & Missed panel & The model does not describe one or more panels of the chart. \\
Diagram & No-label silence & The model writes only ``no label'' for a figure with no printed text. \\
Diagram & Stops at one subject & The model names one part of the diagram and stops, instead of covering every part. \\
Diagram & Unnamed subject & The model does not say what a face or eye diagram shows, when it has no printed labels. \\
\hline
\end{tabular}
\label{tab:defects}
\end{table}

\begin{figure}[!t]
\centering
\includegraphics[width=1.00\columnwidth]{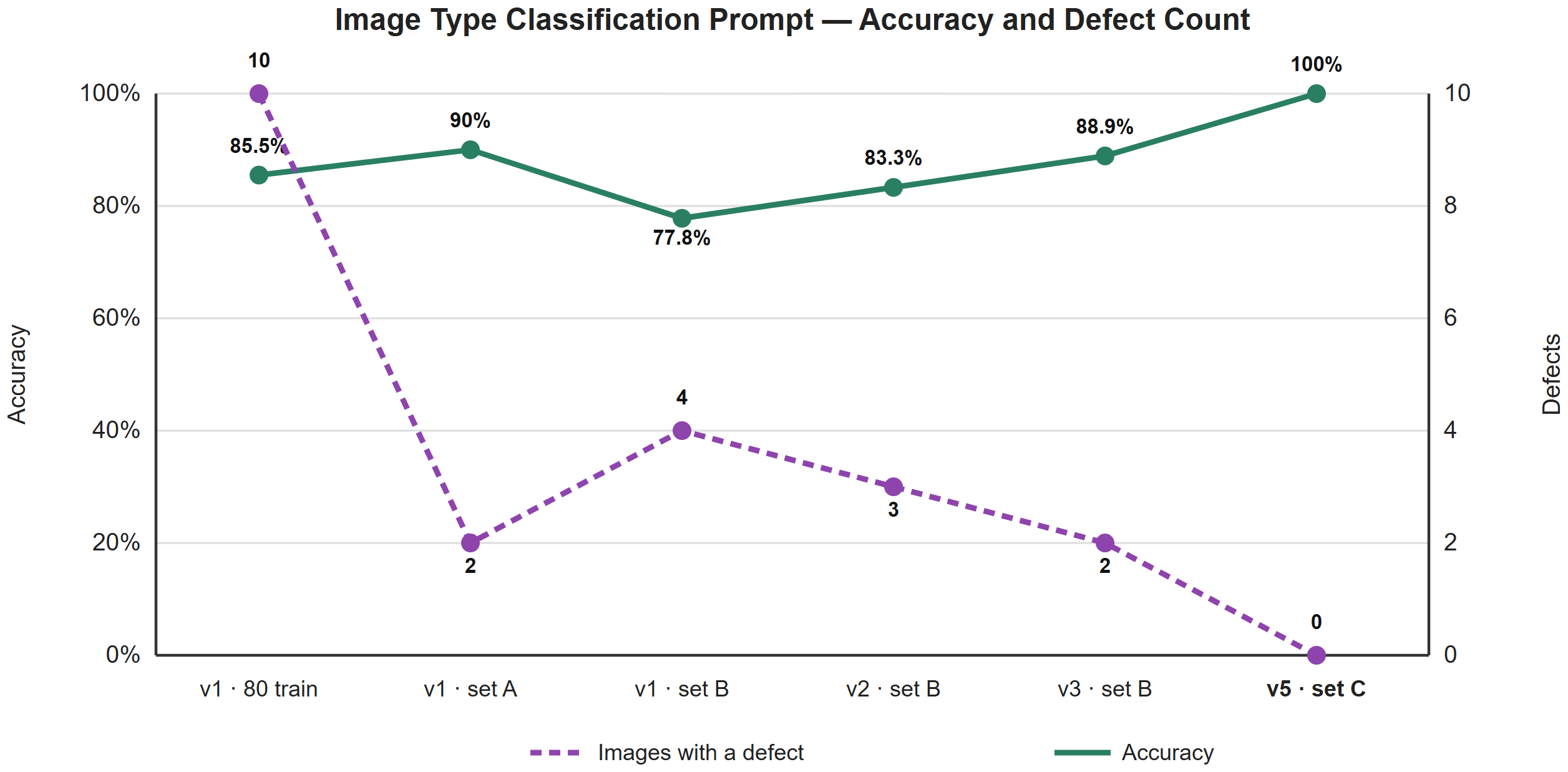}
\caption{Accuracy and defect count of each classification prompt version, by test set.}
\label{fig:8}
\end{figure}

We tested each version against 100 hand-labelled images, and checked each answer against the authors' own label for that image. v1 had 4 classes: table, chart, diagram, photo. We tested it on 69 training images and a 20-image check set. It had the first two defects of Table~\ref{tab:defects}, and also called a diagram a photo. v2 added rules against these defects. It still made some wrong table calls and mix-ups on a harder 18-image set. v3 made these rules stricter and fixed the mix-up. v4 added a confusion matrix and heatmap to the chart class. This fixed the missed confusion matrix defect on one image. We did not run v4 on a full set. v5 added the classes screenshot and other. It removed the line that forced every leftover image into table, chart, or diagram. This fixed the forced-match defect: v5 got 20 of 20 on a new, held-out check set. We select v5. Figure~\ref{fig:8} shows the accuracy and defect count of each version.

A chart may carry more numbers than a diagram does. For this reason, we worked on the chart prompt next and spent more effort on it.

\subsection{The chart prompt}

\begin{figure}[!t]
\centering
\includegraphics[width=1.00\columnwidth]{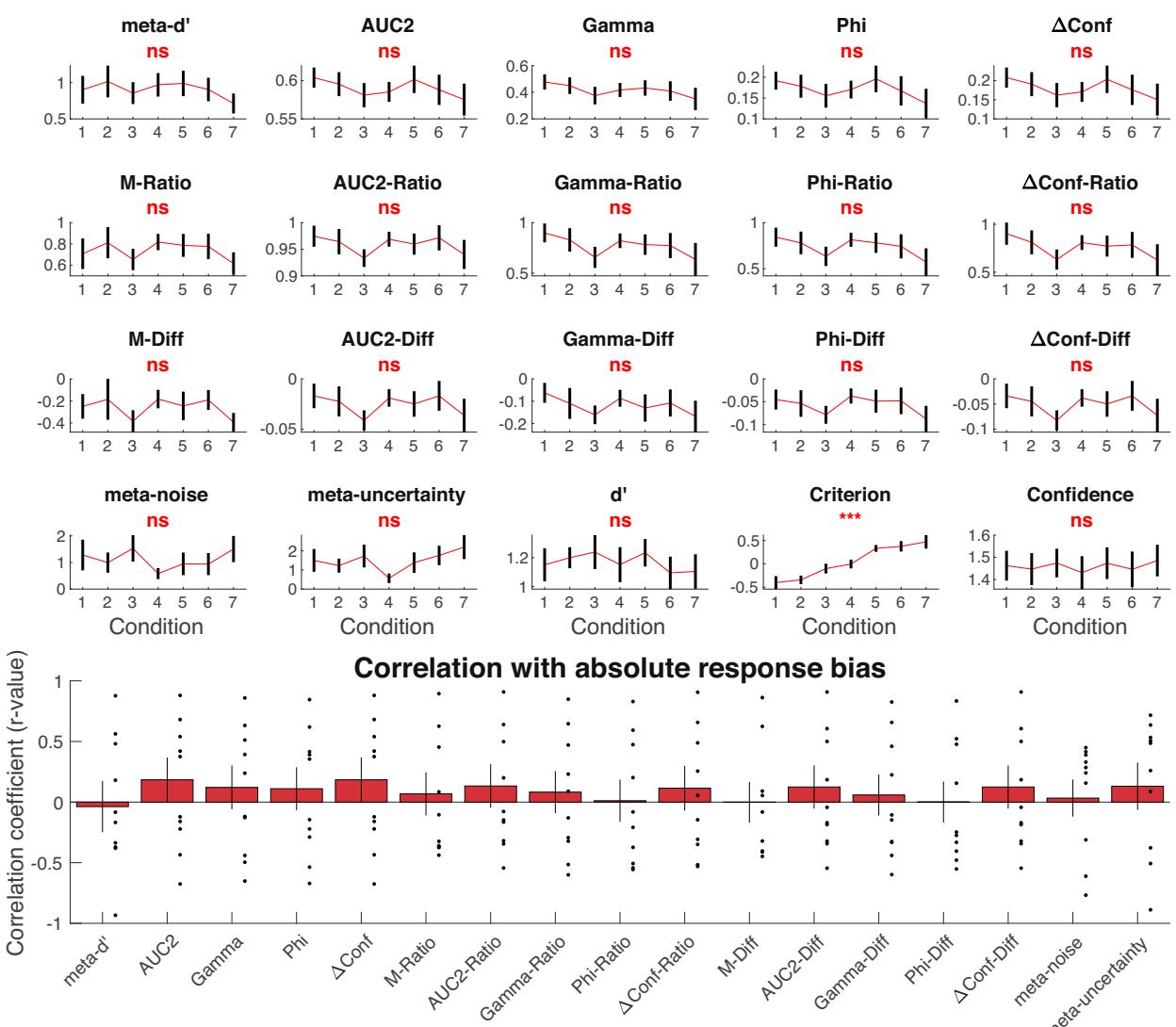}
\caption{The hard image used to test each chart prompt version: a 21-panel figure with 20 line charts and one boxplot. Source: Rahnev [20].}
\label{fig:9}
\end{figure}

We changed the chart prompt six times, to raise the F1 score and remove a defect each time. Table~\ref{tab:defects} names each defect. We tested each version on one hard image: a 21-panel figure with 20-line charts and one boxplot (Figure~\ref{fig:9}). We compared the answer for each panel with the real published panel, and marked a defect where they differ.

\begin{figure}[!t]
\centering
\includegraphics[width=1.00\columnwidth]{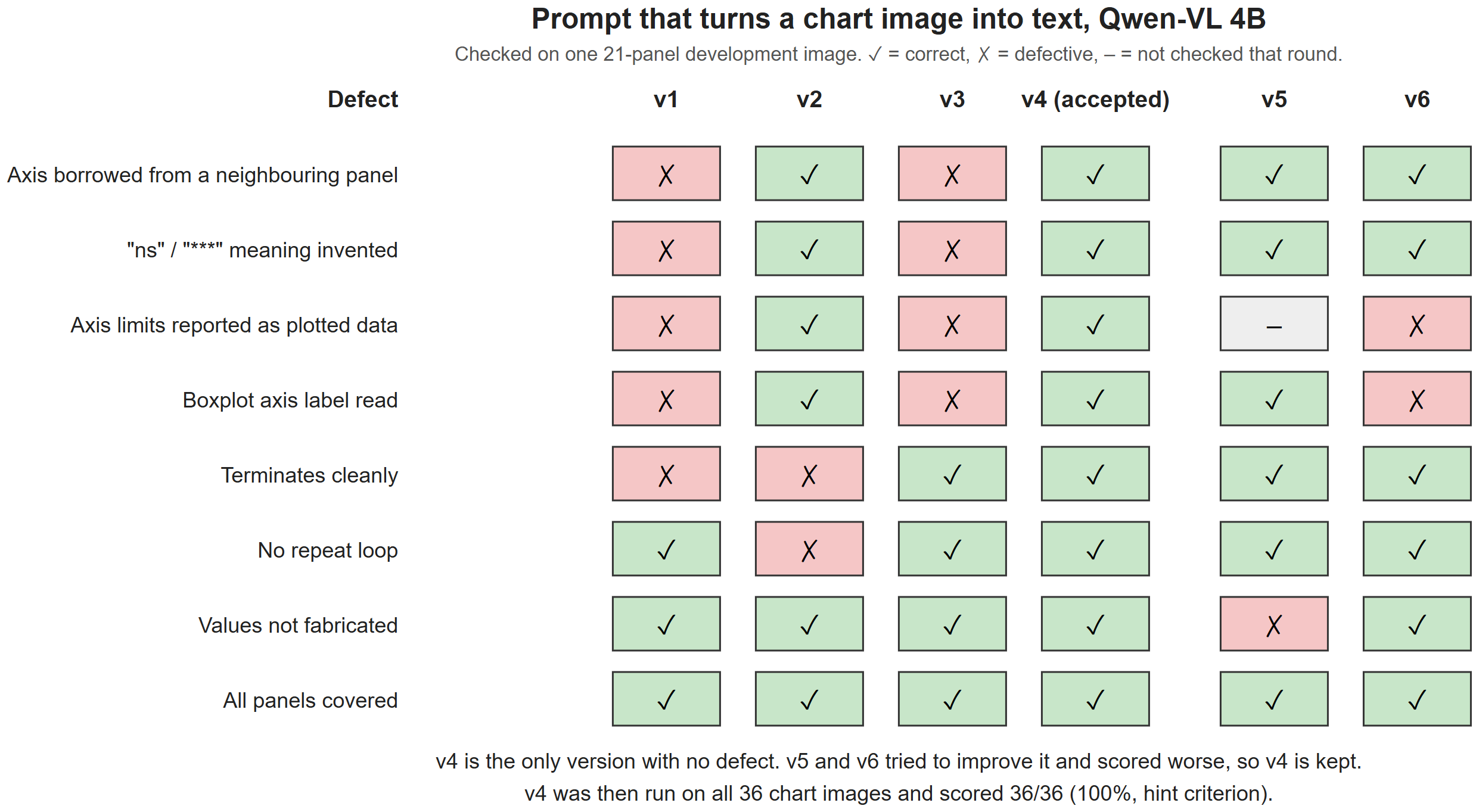}
\caption{The defect matrix of each chart prompt version, on the one development image.}
\label{fig:10}
\end{figure}
\begin{figure}[!t]
\centering
\includegraphics[width=1.00\columnwidth]{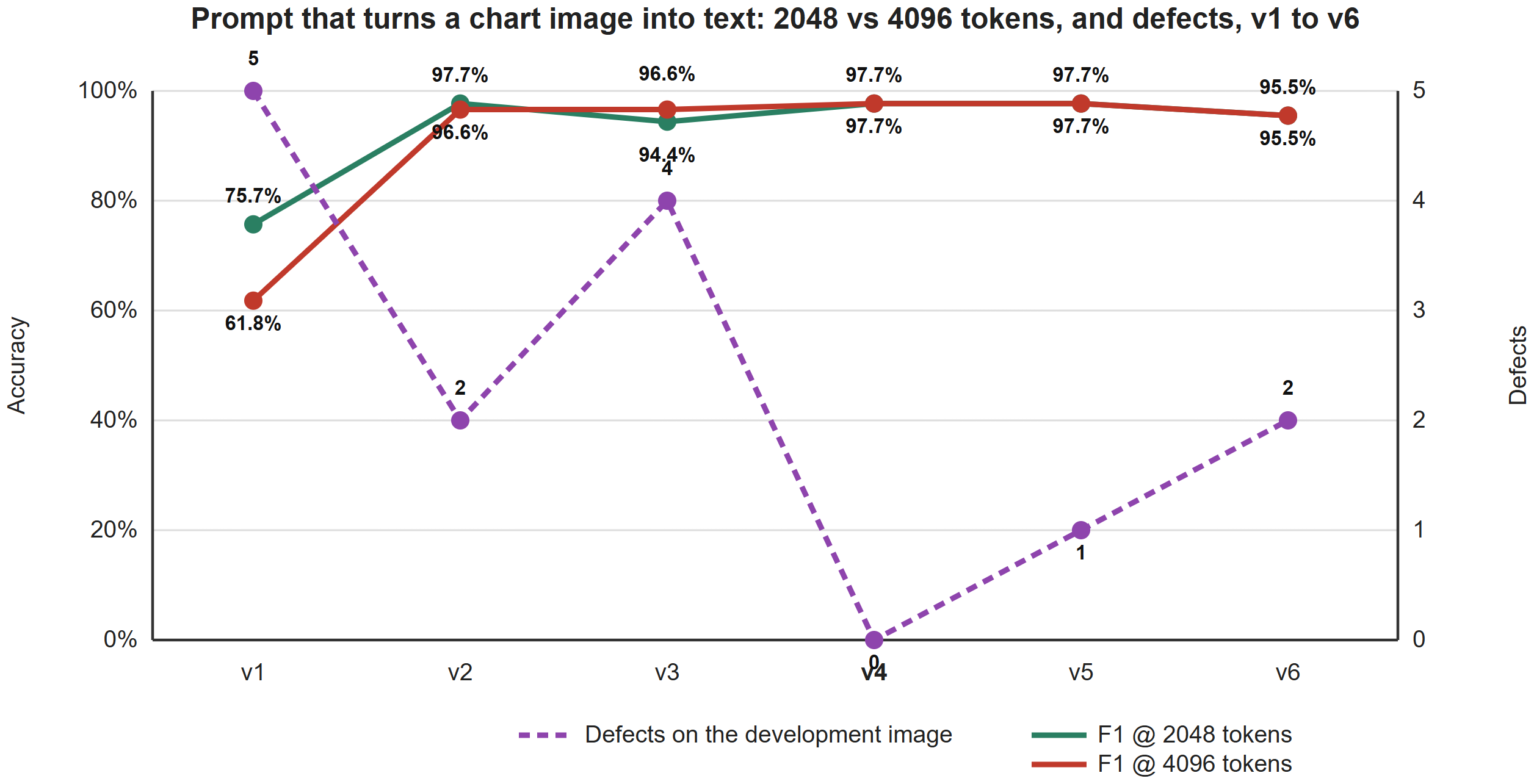}
\caption{F1 at output length limit 2048 vs 4096, with the defect count on the right axis, v1 to v6.}
\label{fig:11}
\end{figure}


v1 had the worst F1 on the full validation set. v2 fixed all four of v1's labelling defects. But its closing instruction made the model repeat itself. v3 removed v2's rules and stopped cleanly, but v1's four defects came back. v4 kept v2's four labelling rules and added a clean-stop instruction, so it fixed every defect at once. v5 and v6 tried to improve v4 and made it worse. v5's added rule made the model invent values. v6's changed rule lost the boxplot-label fix. We first set Qwen3.5-4B's token limit to 2048. On the full validation set of 36 chart images, v2, v4, and v5 then tie on F1. v4 is the only one of the three with no defects, so we select v4 on the defect count. Figure~\ref{fig:10} shows the defects, and Figure~\ref{fig:11} shows the F1 of each version.

We also ran every version on all 36 images at a token limit of 4096. The F1 of v4 does not change. A higher limit does not help any version beat v4. Figure \ref{fig:11} compares the two limits.

\begin{figure}[!t]
\centering
\includegraphics[width=1.00\columnwidth]{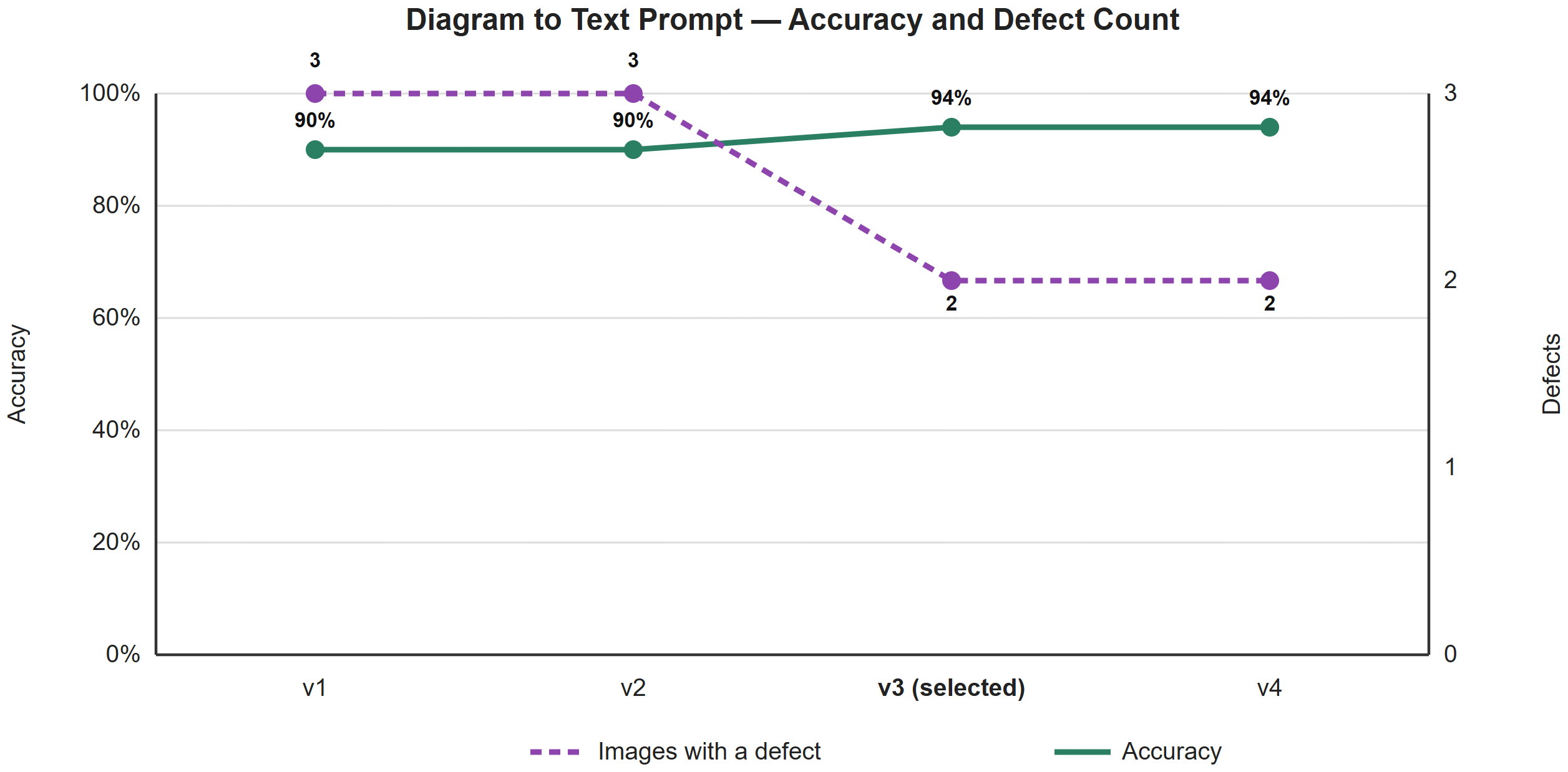}
\caption{Accuracy and defect count of each diagram prompt version, 31-image validation set.}
\label{fig:12}
\end{figure}

\subsection{The diagram prompt}

We changed the diagram prompt four times, using the structure of the chart prompt, with the same goal as before. Table~\ref{tab:defects} names each defect. We tested each version on all 31 diagram images, against the hand-checked description of each.

v1 scored 28 of 31 diagrams correct. The three failed images all had the same defect: no-label silence. v2 added a rule to open with the diagram's subject. It fixed two of these images. But the new rule caused the defect ``stops at one subject'', so two other diagrams lost detail. v2 scored 28 of 31 again. v3 replaced that rule with a rule to cover every part before stopping. This fixed both new failures. v3 scored 29 of 31, with two failed images left. Both are diagrams of a face or an eye, drawn only as dots and with no printed label. The model does not name what they show. Figure~\ref{fig:12} shows the accuracy and defect count of each version. v4 added a named example to fix these two images. The accuracy stayed at 29 of 31. One failure became worse: the model named the wrong subject. v3 and v4 tie on accuracy and defect count, but v4 names the wrong subject in one image. We reject v4 and select v3. We ran the diagram prompt at a token limit of 4096 only. Unlike the chart prompt, we made no comparison against a lower limit.

\subsection{Tables}

The MinerU parser parses most tables on its own while it reads the page. It turns each one into a <table> tag, with no model and no prompt. A table reaches our classification prompt only when the page holds it as a plain picture. The classification prompt names it table. We send it to the same MinerU parser, so every table ends as the same <table> tag. If MinerU returns an empty description, the table may be too complex or may contain other kinds of content. We then process it with Qwen3.5-4B.

We tested this choice first. Table~\ref{tab:qwentables} names what went wrong with Qwen3.5-4B.

\begin{table}[!t]
\caption{Problems of Qwen3.5-4B on tables.}
\centering
\footnotesize
\begin{tabular}{p{0.30\columnwidth}p{0.60\columnwidth}}
\hline
\textbf{Problem} & \textbf{What it means} \\
\hline
Ran out of room & The model stops partway through the table, because the answer is too long for its token limit. \\
Wrong grid & The model draws the rows and columns in a different shape from the real table, when cells span more than one row or column. \\
Repeats itself & The model writes the same row over and over, instead of stopping. \\
\hline
\end{tabular}
\label{tab:qwentables}
\end{table}

We drew 30 tables from the validation split of PubTabNet [21], a public set of 9,115 tables from PubMed Central articles. Each table comes with its correct answer, as an HTML table. We split them into 24 hard tables and 6 easy control tables. Each hard table has a merged cell or a header that spans two or more rows. The easy tables have no merged cells. The hard tables are of four kinds, six tables each: a deep header, many cells that span rows, a large grid, and many cells that span columns. We tested two parsers on these tables and scored each answer cell by cell against the real table.

\begin{figure}[!t]
\centering
\includegraphics[width=1.00\columnwidth]{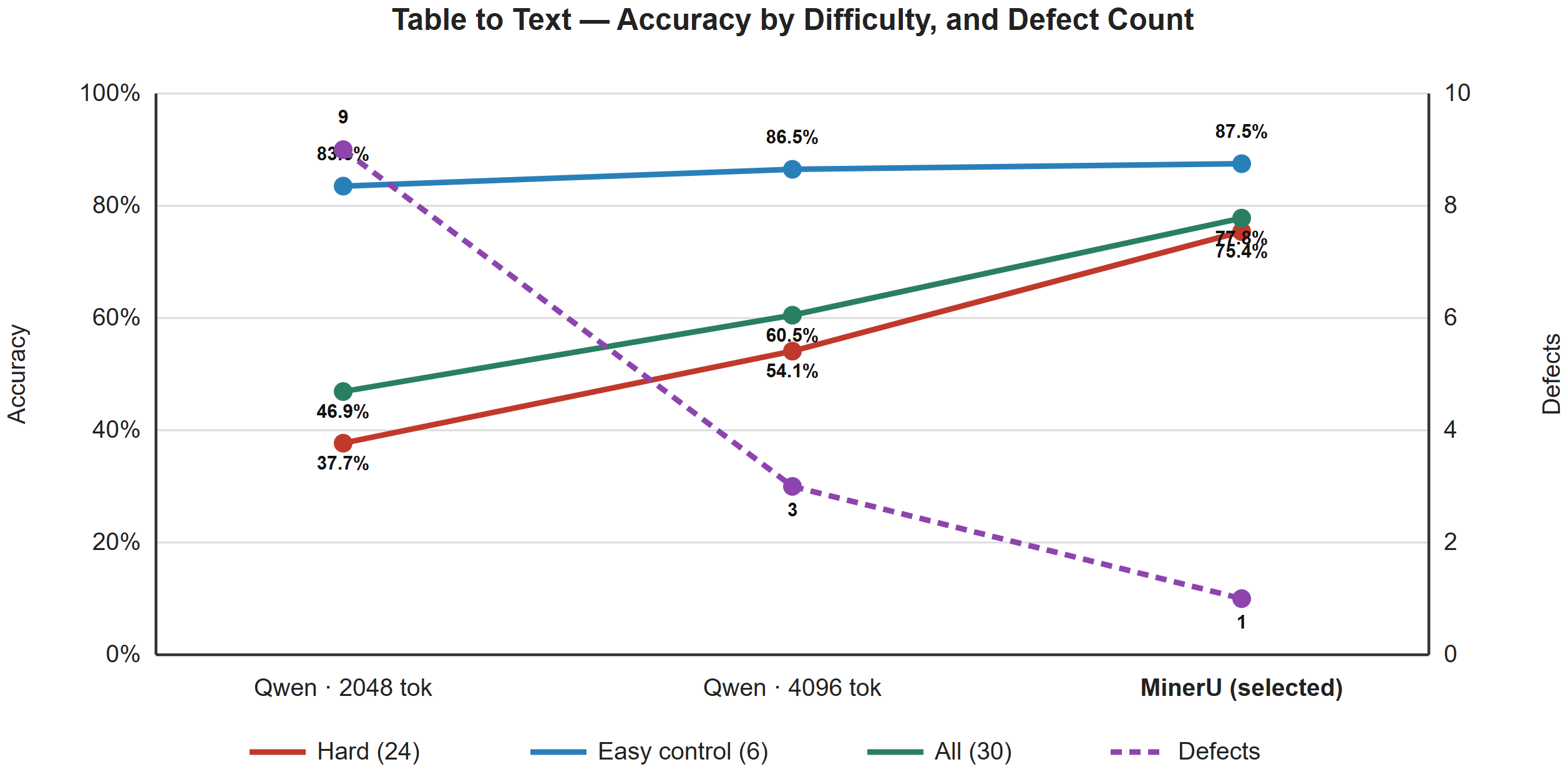}
\caption{Accuracy, by table difficulty, and defect count, Qwen3.5-4B at two token limits against MinerU, 30-table validation set.}
\label{fig:13}
\end{figure}

Qwen3.5-4B, with a token limit of 2048, had 9 tables that failed completely. Most ran out of room: the limit, not the model, was the problem. Raising the limit to 4096 left 3 tables that failed completely. More room solved the token-limit problem. But it did not fix how Qwen3.5-4B reads a table's grid. The hard tables stayed far behind the easy tables. The model kept drawing the wrong grid: it merged or split cells that the real table does not. MinerU, with no prompt and no token limit, scored best on the hard tables and overall. Only 1 table failed completely. We select MinerU. It reads the grid correctly far more often, and it cannot run out of room. Figure \ref{fig:13} shows these scores.

\textbf{Limitation.} MinerU still gets 1 table in 30 wrong. Its score on hard tables, 75.4\%, is well below its score on easy tables, 87.5\%. A table parser built and trained only for tables may read merged cells and multi-row headers better, and may run faster. This is future work.

\section{The XML image block}

Each image in the text carries one <image> block, with three tags: <type> (the image's class — chart, diagram, photo, or table), <url> (the link to the real image file), and <text> (a short description written by the model). An agent reads <text> first: the cheapest level, no image file needed. If this text does not answer the agent's question, the agent can ask the model for more detail from the same description, still with no image file. If the text still falls short, the agent uses <url> to load the real image, and asks the vision-language model a new question straight on that image. This gives the agent three levels of depth — the short text, more text from the same description, then the real image — and the agent pays for the slower, deeper level only when the level before it was not enough.

\section{THE MCP SERVER and the SKILL}

So that any agent can use this parser, we created an MCP server, delivered with all the models as a Docker network. The MCP server exposes three services used for document parsing: MinerU, which reads a document's layout; quantized Popo, which repairs its heading structure; and Qwen3.5-4B, which classifies and describes each image. Each service has its own tools and its own resources on how to use those tools in the MCP layer. An agent reaches these services through one skill, through the MCP server, and through the tools and resources the server exposes. The skill, convert-pdf, connects the agent to the MCP server. A tool calls one service alone. The pipeline calls many services in a fixed order. A resource tells the agent how to use a tool or the pipeline. Each service runs from its own Docker image, for one job only, and the system removes it after the job. Only one service holds the GPU at a time. The server also holds three services that this paper does not test: WhisperX (audio and video to text), Embed (one vector per text), and Prep (a web page to PDF).

A full document, PDF to MD, is not a single service call. It is one fixed pipeline already shown in Figure 1 at the start of this paper. Each tool and the pipeline have their own resource file, stating what the tool needs and returns. Along with the MCP server, we designed a skill (for Claude Code and Cowork) that gives the agent instructions and one way to use the whole system. The skill can be converted for other agents that need this PDF-to-Markdown harness.

\section{Evaluation}

\begin{figure}[!t]
\centering
\includegraphics[width=1.00\columnwidth]{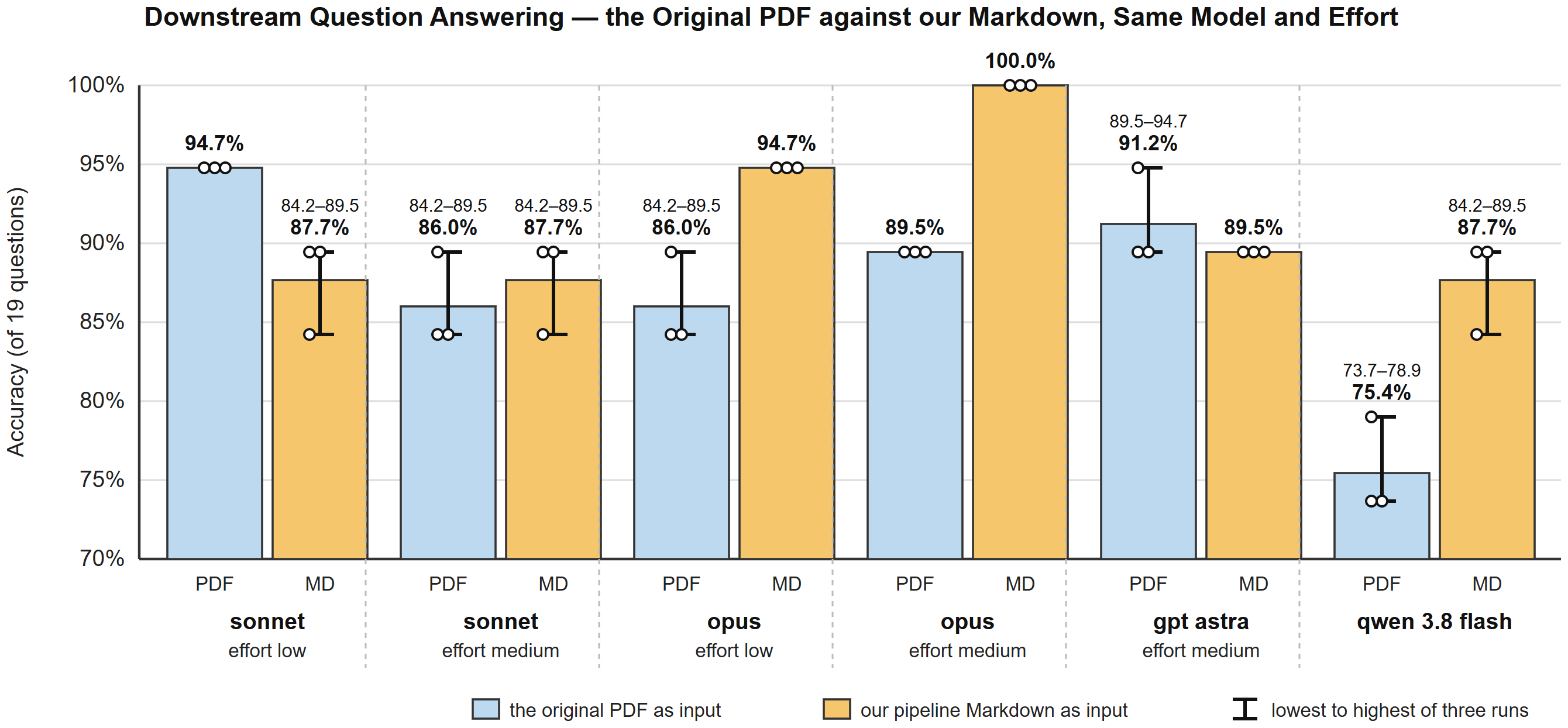}
\caption{The same model and effort, reading the PDF against reading the parsed Markdown, six matched pairs, three-run range shown.}
\label{fig:15}
\end{figure}

This test asks one question. Which input lets a model answer a question about a paper’s figures better? The two inputs are the original PDF and the parsed Markdown. Five models answer nineteen questions whose answer exists only inside a figure. Table~\ref{tab:setups} lists the models and their setups. One model, with one input and one reasoning effort, is one setup. There are fourteen setups in total: six read the PDF, and eight read the Markdown. Each setup ran three times, to separate a real difference from chance. This gives 798 graded answers: 14 setups $\times$ 3 runs $\times$ 19 questions. Figure \ref{fig:15} compares the same model, at the same reasoning effort, on the two inputs PDF and Markdown. The parsed Markdown wins four of the six matched pairs. One model, Qwen 3.8 Flash [22], is open-weight. Its weights are public, so it can run locally. This model gains 12.3 points on the Markdown over the PDF. This closes most of its gap to the closed, frontier models.

\begin{table}[!t]
\caption{The fourteen test setups. API: the model ran through the OpenRouter API [23], with no file access. Agent: the model ran as an agent through the Claude API, with a tool to read files. Image fallback: the agent could also open the original figure image.}
\label{tab:setups}
\centering
\footnotesize
\begin{tabular}{llllc}
\hline
Model & Input & Effort & Access & Image fallback \\
\hline
Sonnet 5.5 & PDF & low & Agent & no \\
Sonnet 5.5 & PDF & medium & Agent & no \\
Opus 5.5 [24] & PDF & low & Agent & no \\
Opus 5.5 & PDF & medium & Agent & no \\
GPT-6 Luna [25] & PDF & medium & API & no \\
Qwen 3.8 Flash & PDF & reasoning on & API & no \\
Sonnet 5.5 & Markdown & low & Agent & yes \\
Sonnet 5.5 & Markdown & medium & Agent & yes \\
Sonnet 5.5 & Markdown & high & Agent & yes \\
Opus 5.5 & Markdown & low & Agent & yes \\
Opus 5.5 & Markdown & medium & Agent & yes \\
Haiku 5.5 [26] & Markdown & low & Agent & yes \\
GPT-6 Luna & Markdown & medium & API & no \\
Qwen 3.8 Flash & Markdown & reasoning on & API & no \\
\hline
\end{tabular}
\end{table}

\begin{figure}[!t]
\centering
\includegraphics[width=1.00\columnwidth]{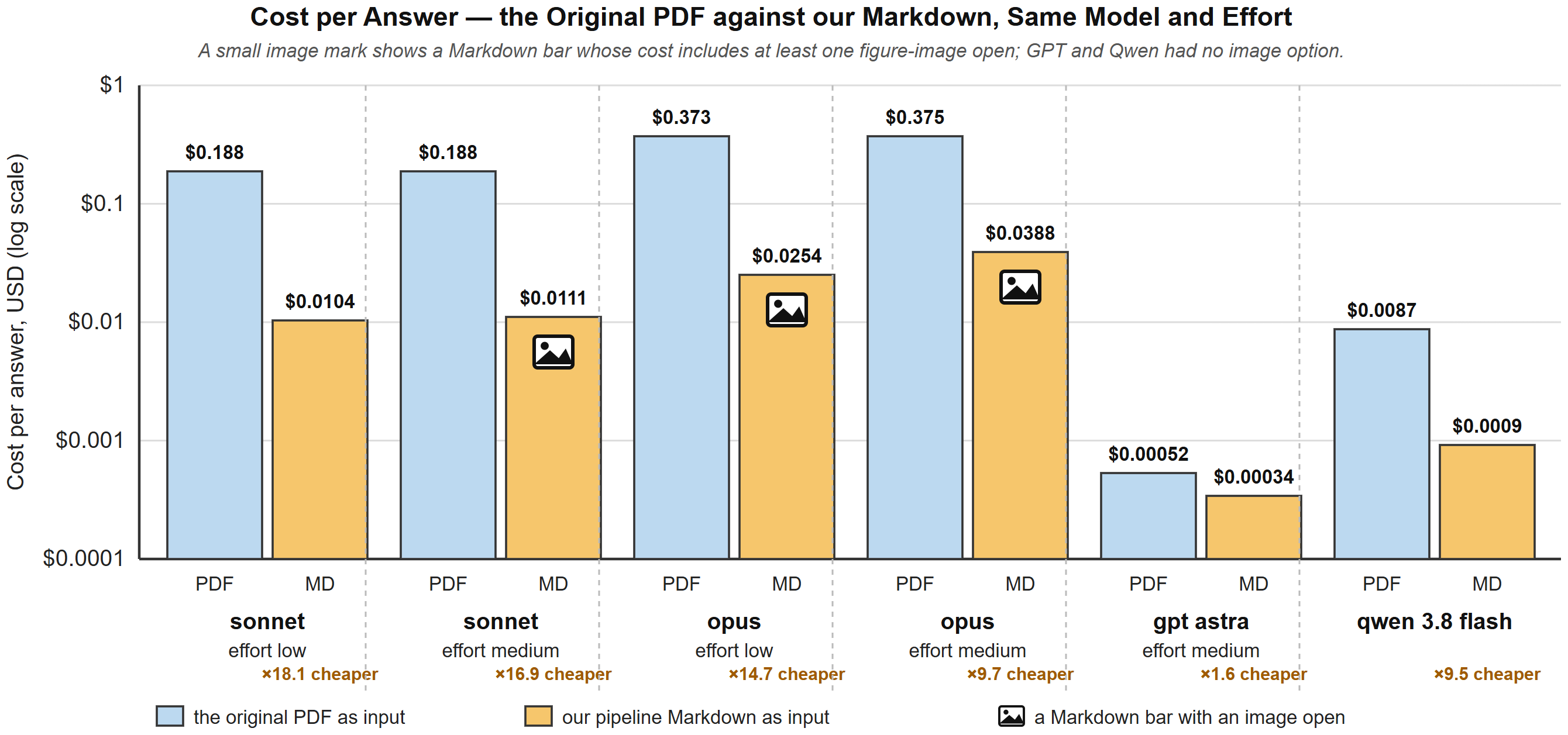}
\caption{The dollar cost of one answer, the PDF against the parsed Markdown, same six pairs, log scale.}
\label{fig:16}
\end{figure}

Figure \ref{fig:16} compares the cost of the same six pairs, one answer at a time. The parsed Markdown costs less than the PDF in every pair, from 1.6 times cheaper to 18.1 times cheaper. A small image mark on a bar means that, in that setup, the agent could also open the real figure images if it needed them. The cost includes every extra call a model made to open a figure image.

\begin{figure*}[!t]
\centering
\begin{minipage}[t]{0.64\textwidth}
\vspace{0pt}
\centering
\includegraphics[width=1.00\linewidth]{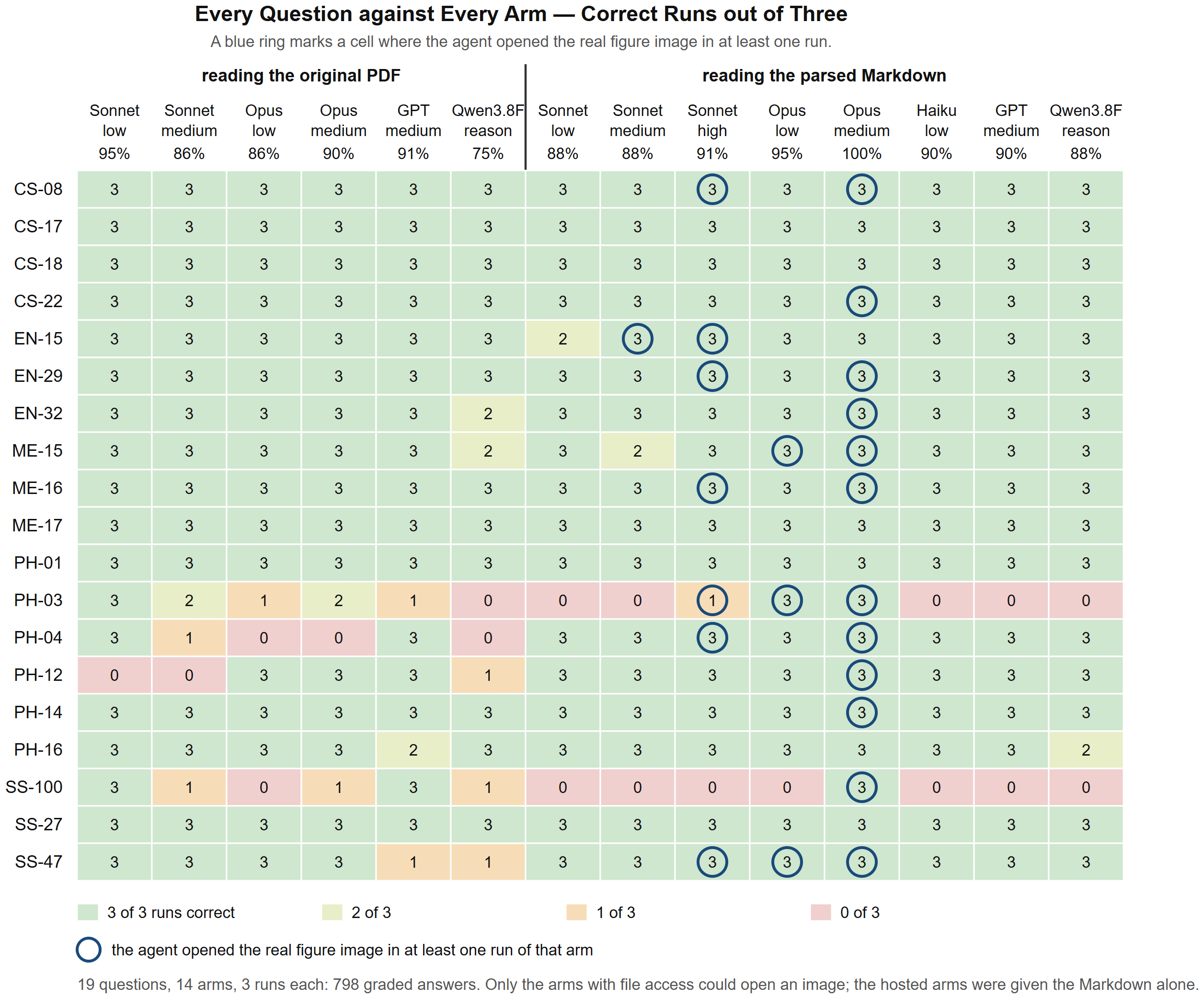}
\caption{All nineteen questions against all fourteen model-and-input setups; each cell counts the correct runs out of three, ringed where a figure image was opened.}
\label{fig:17}
\end{minipage}\hfill
\begin{minipage}[t]{0.34\textwidth}
\vspace{0pt}
\centering
\includegraphics[width=1.00\linewidth]{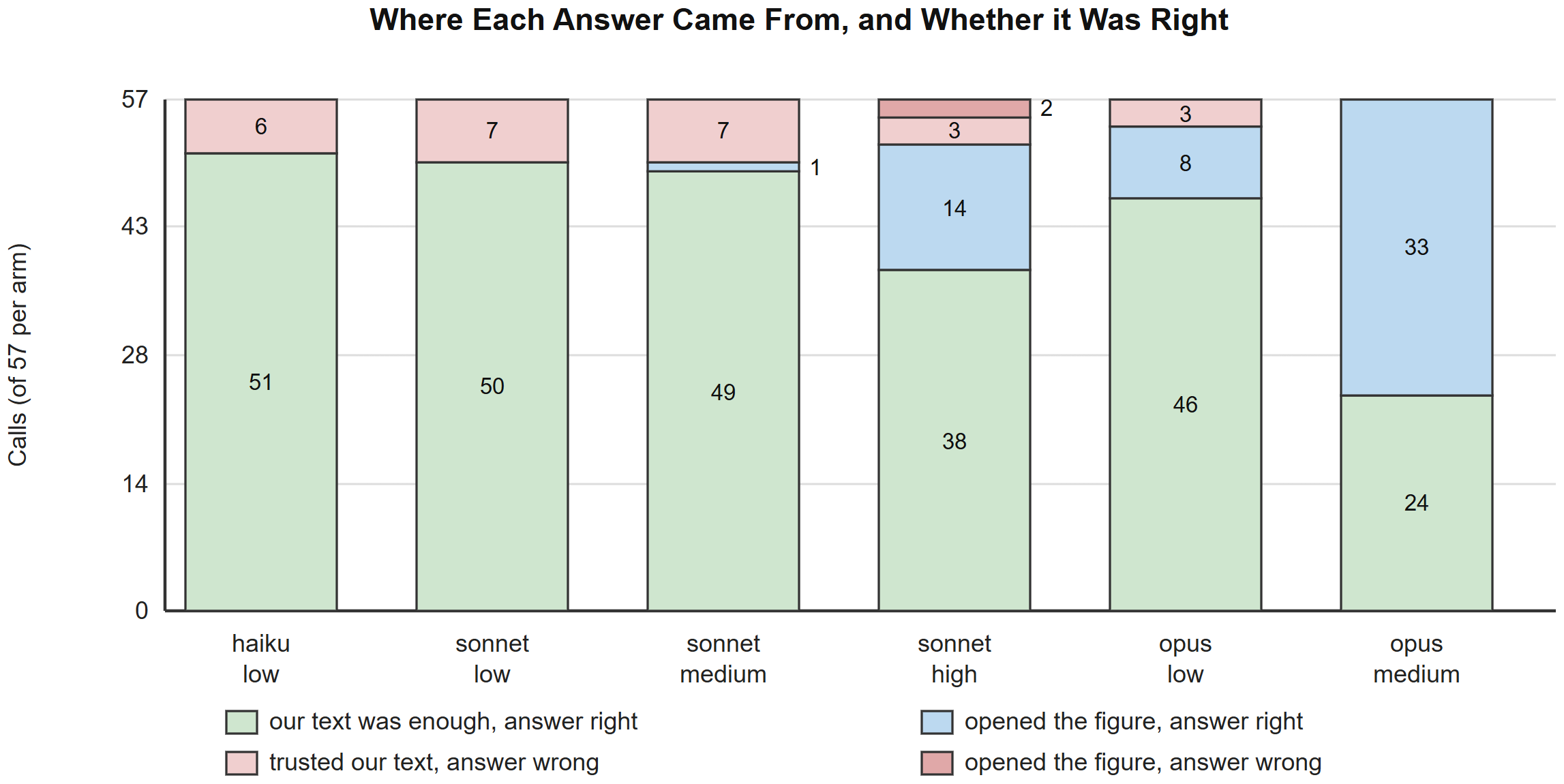}
\caption{Every answer from the models with an image fallback, split by whether the text alone was enough and whether the answer was right.}
\label{fig:18}
\vspace{4pt}
\includegraphics[width=1.00\linewidth]{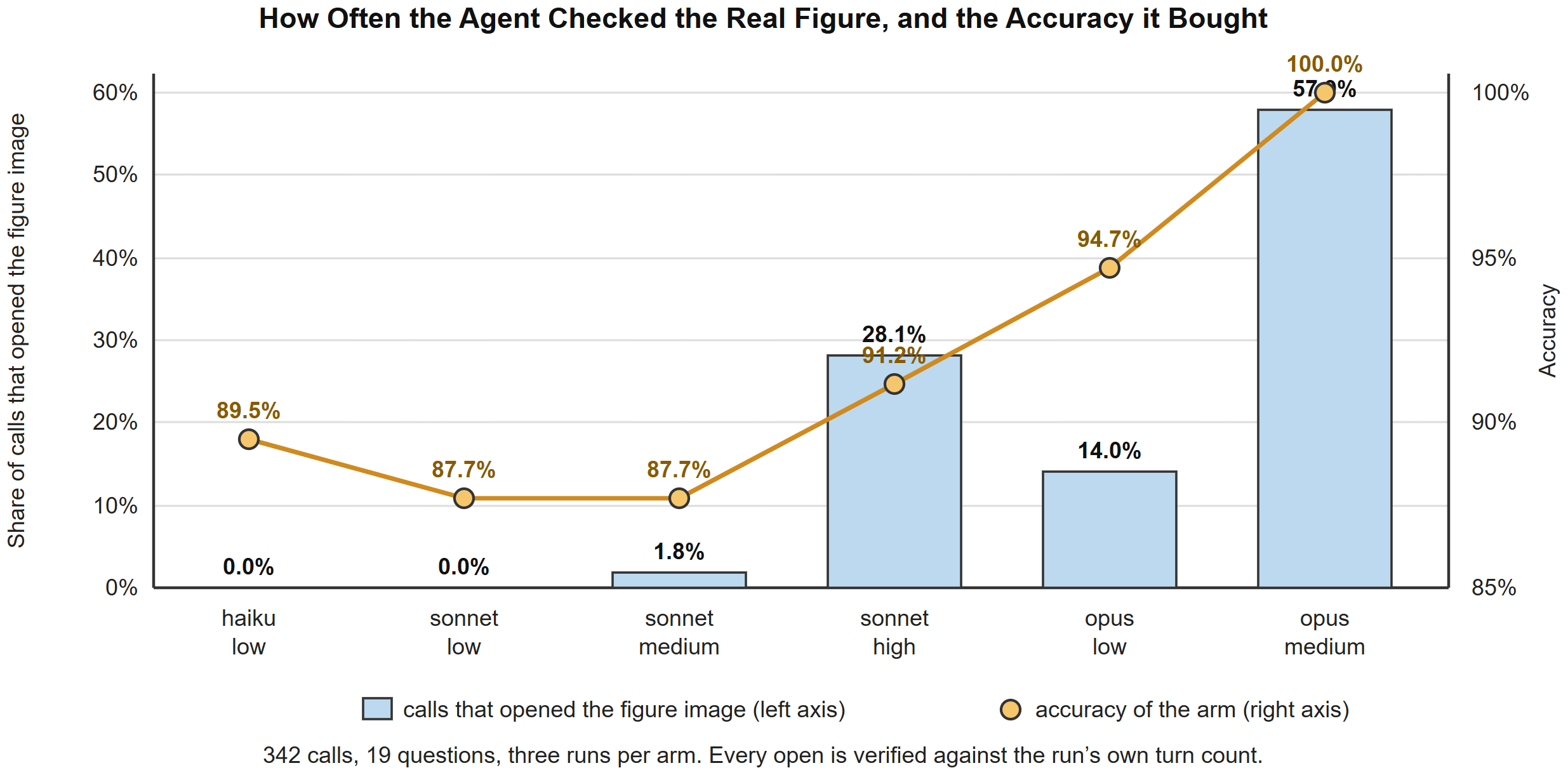}
\caption{How often each setup opened the original figure image, against that setup's accuracy.}
\label{fig:19}
\end{minipage}
\end{figure*}

Figure \ref{fig:17} shows every question against every setup. Each row is one of the 19 questions, and each column is one of the 14 setups of Table~\ref{tab:setups}: six that read the PDF on the left, and eight that read the Markdown on the right. Nine of the nineteen questions are correct in every setup. Each asks for a number already printed in the figure. These nine questions do not separate the two inputs. Two questions fail only on a PDF setup: a model reading the rendered page misjudges an unlabelled plot that the pipeline's own text states correctly. Two questions fail only on a Markdown setup. Both have one cause. The vision-language step wrote the wrong value for an unlabelled chart. In one case it read a stacked chart's top instead of one band's own height. In the other case it read the wrong curve's peak. No other question fails on the Markdown alone. These two wrong descriptions are the only errors that the Markdown adds over the PDF.

Three closed models (Sonnet, Opus and Haiku in Table~\ref{tab:setups}) could also open the original figure image, in their Markdown setups, when the parsed text did not hold enough. The prompt gave one fixed rule. Read the text first. Open the image only if the text still lacks the value. Figure \ref{fig:18} shows the result over the 342 answers from these three models' Markdown setups. The text alone was right in 258 of the 342 answers, with no image opened. An image was opened in 58 of the 342 answers, and 56 of those 58 opens were correct, recovering both of the two failures named above. Figure \ref{fig:19} shows that a model's own choice to open an image, not the prompt, decides how often it opens. The open rate ranges from 0 \% to 57.9 \% across the six setups with an image fallback, at the same prompt. A higher open rate goes with a higher score.

With the image fallback available, one setup answers all nineteen questions correctly in every run. The pipeline's remaining errors are two known, unlabelled-plot misreadings, not a layout or text-extraction fault.

\section{Performance}

This section shows the cost of the method, its bottleneck, and the future work that could lower it. The pipeline converted 30 papers in 10,492.5 s, or 2.91 h, in total. The mean time per paper was 349.8 s, with a median of 254.3 s. The machine was one laptop, an MSI-L. It has an Intel Core Ultra 7 255HX CPU and 32.0 GB of RAM. Its GPU is an RTX 5070 Ti Laptop GPU, with 12 GB of VRAM.

\begin{figure}[!t]
\centering
\includegraphics[width=1.00\columnwidth]{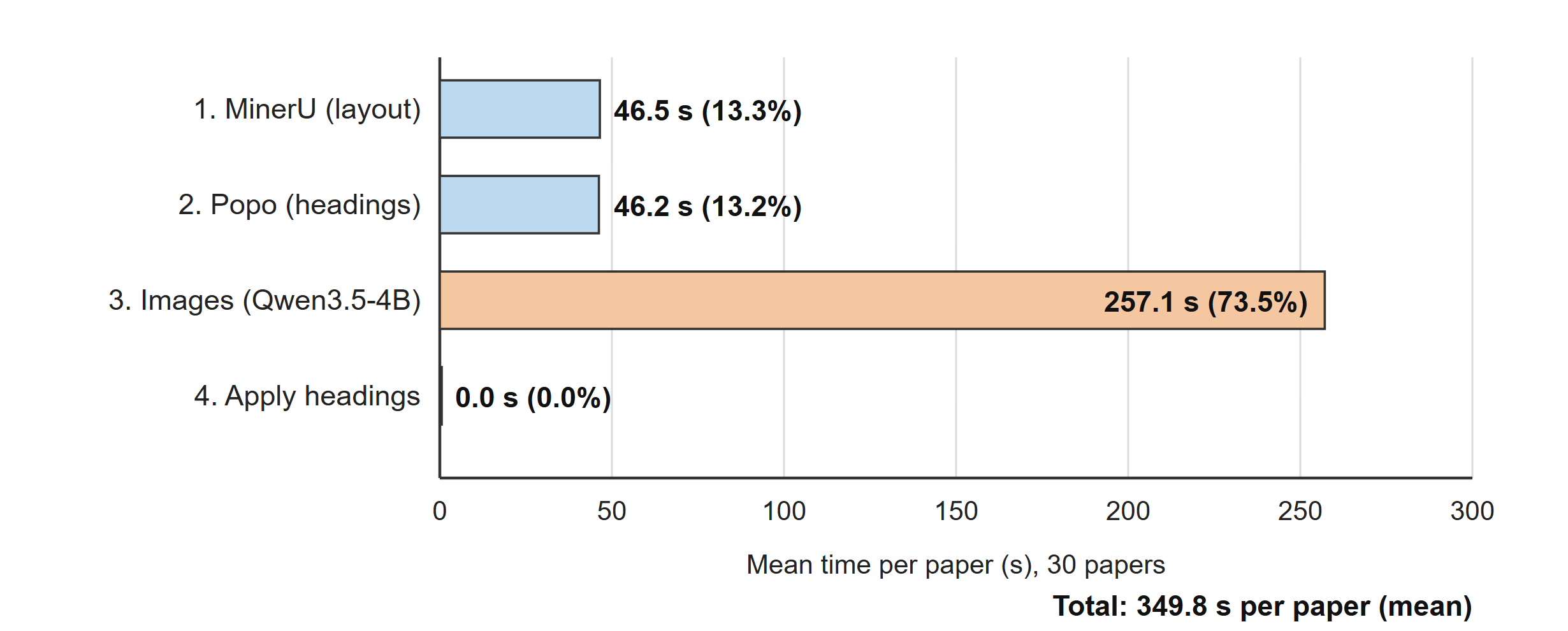}
\caption{Mean time per paper of each pipeline stage, over the 30 papers, on one laptop GPU.}
\label{fig:20}
\end{figure}

The image stage uses the Qwen3.5-4B vision-language model to describe every figure. It took 73.5\% of the total time. The two text stages, MinerU and Popo, took the other 26.5\%. Figure \ref{fig:20} shows the mean time of each stage.

The bottleneck is the 4B vision-language model, not the text conversion. It reads each figure with a 4K-token limit. Two paths can speed up this stage in future work. One path runs a larger vision-language model on a larger GPU, which may read figures faster, or more accurately. The other path uses a smaller model built only for one figure type. Three such models exist for tables: TeleOCR [27], TRivia-3B [28], and UniTable [29]. Similar models for charts can also be tested. Future work tests whether these models raise accuracy while the pipeline still runs on one local GPU.

\section{Limitations}

This paper tests one vision-language model, Qwen3.5-4B, not several. The parser comparison tests a set of parsers, not every parser available. The 30-paper dataset comes from diverse sources, not one controlled sampling method, and covers only five fields, all research papers. The downstream evaluation's images come only from these same 30 papers. No model built for one image type alone was tested. The downstream evaluation tests only Claude, GPT, and Qwen 3.8 Flash models, mostly at low or medium reasoning effort, with high effort tested for one model only.

\section{Conclusion}

The downstream test compared the original PDF and the parsed Markdown as input. The parsed Markdown scored higher than the PDF in four of six matched pairs. The largest gain was for the one open-weight model, Qwen 3.8 Flash. Its score rose from 75.4\% on the PDF to 87.7\% on the Markdown, a gain of 12.3 points. An open-weight model can run on a local machine, so this result supports a fully local solution for private data. This is the paper's main result. A clear, structured Markdown file can raise a model's own accuracy on figure-based questions, not only keep the question answerable.

The pipeline keeps private data on one local machine. Qwen3.5-4B classifies and describes each figure, and MinerU parses each table. One local MCP server runs the whole pipeline, on one laptop, with one GPU and 12 GB of memory. No document ever leaves this machine.

The parsed Markdown also costs less to read than the original PDF, from 1.6 times to 18.1 times less, for the same questions. This saving grows with the number of documents processed.

The <image> tag holds each figure's type, link, and description in one place, not in free text. This structure lets the output feed a knowledge graph or an ontology later. This pipeline is the first part of a larger, local harness system.

In their Markdown setups, three models (Sonnet, Opus and Haiku) had the original figure image as a fallback. They gave 342 answers from the parsed Markdown. Text alone was enough for 258 of these answers. An image was opened for only 58 answers, and 56 of those 58 opens were correct. The model mostly chose by itself when the text was enough and when it needed the image.

\section*{Acknowledgments}

This work is supported by the MPhil program ``Advanced Technologies in Informatics and Computers'', which is hosted by the Department of Informatics, Democritus University of Thrace, Kavala, Greece. The authors used an AI assistant, Claude (Anthropic), in this research, to help write the paper in a clearer form. A human was in the loop at every step: a human gave each instruction, observed the work, and evaluated each result. The AI assistant asked the human for every decision, and the human made each decision. The authors checked all text, numbers, and results, and take full responsibility for the content of this paper.


\begin{thebibliography}{00}
\bibitem{b1} Q. Zhang, B. Wang, V. S.-J. Huang, J. Zhang, Z. Wang, H. Liang, C. He, and W. Zhang, ``Document parsing unveiled: Techniques, challenges, and prospects for structured information extraction,'' arXiv:2410.21169, (2024).
\bibitem{b2} Y. Ding, S. C. Han, J. Lee, and E. Hovy, ``Deep learning based visually rich document content understanding: A survey,'' Artificial Intelligence Review, Springer, DOI: 10.1007/s10462-025-11477-3, (2026).
\bibitem{b3} P. Lewis, E. Perez, A. Piktus, F. Petroni, V. Karpukhin, N. Goyal, H. Küttler, M. Lewis, W. Yih, T. Rocktäschel, S. Riedel, and D. Kiela, ``Retrieval-augmented generation for knowledge-intensive NLP tasks,'' Advances in Neural Information Processing Systems (NeurIPS), vol. 33, pp. 9459-9474, (2020).
\bibitem{b4} K. Ordoumpozanis, M. Konstantakis, J. Aliprantis, A. Angeioplastis, D. Varsamis, G. Caridakis, and A. Tsimpiris, ``Playing telephone with AI: How uninformed use of AI tools may distort educational outcomes,'' Proc. 2026 International Conference on Advances in Artificial Intelligence and Machine Learning (AAIML), 936-944, (2026).
\bibitem{b5} I. Shumailov, Z. Shumaylov, Y. Zhao, N. Papernot, R. Anderson, and Y. Gal, ``AI models collapse when trained on recursively generated data,'' Nature 631, 755--759, DOI: 10.1038/s41586-024-07566-y, (2024).
\bibitem{b6} I. C. Wiest et al., ``Privacy-preserving large language models for structured medical information retrieval,'' npj Digital Medicine, vol. 7, article 257, DOI: 10.1038/s41746-024-01233-2, (2024).
\bibitem{b7} D. Jiang, Y. Li, G. Li, and B. Li, ``MAGMA: A multi-graph based agentic memory architecture for AI agents,'' Proceedings of the 64th Annual Meeting of the Association for Computational Linguistics (Volume 1: Long Papers), pp. 36848-36865, (2026).
\bibitem{b8} Y. Hu, Y. Wang, and J. McAuley, ``Evaluating memory in LLM agents via incremental multi-turn interactions,'' International Conference on Learning Representations (ICLR), (2026).
\bibitem{b9} Y. Xu, M. Li, L. Cui, S. Huang, F. Wei, and M. Zhou, ``LayoutLM: Pre-training of text and layout for document image understanding,'' Proceedings of the 26th ACM SIGKDD International Conference on Knowledge Discovery \& Data Mining, pp. 1192-1200, (2020).
\bibitem{b10} B. Xu, Z. Miao, X. Zhou, Y. Lin, Z. Tang, X. Zhao, F. Wu, C. Tan, F. Wu, B. Wang, and C. He, ``MinerU-Popo: Universal post-processing model for structured document parsing,'' arXiv:2605.24973, (2026).
\bibitem{b11} K. Ordoumpozanis, ``MinerU-Popo-NF4,'' Hugging Face model repository, (2026).
\bibitem{b12} B. Wang, C. Xu, X. Zhao, L. Ouyang, F. Wu, Z. Zhao, R. Xu, K. Liu, Y. Qu, F. Shang, B. Zhang, L. Wei, Z. Sui, W. Li, B. Shi, Y. Qiao, D. Lin, and C. He, ``MinerU: An open-source solution for precise document content extraction,'' arXiv:2409.18839, (2024).
\bibitem{b13} Qwen Team, ``Qwen3.5: Towards native multimodal agents,'' Qwen blog, (2026).
\bibitem{b14} J. Priem, H. Piwowar, and R. Orr, ``OpenAlex: A fully-open index of scholarly works, authors, venues, institutions, and concepts,'' arXiv:2205.01833, (2022).
\bibitem{b15} Anthropic, ``Claude Sonnet 5.5,'' official model page, \url{https://www.anthropic.com/claude-sonnet-5-5}, (2026).
\bibitem{b16} Artifex Software, Inc., ``PyMuPDF4LLM,'' GitHub repository, \url{https://github.com/pymupdf/pymupdf4llm}, (2024).
\bibitem{b17} LlamaIndex, Inc., ``LiteParse,'' GitHub repository, \url{https://github.com/run-llama/liteparse}, (2026).
\bibitem{b18} N. Livathinos, C. Auer, M. Lysak, A. Nassar, M. Dolfi, P. Vagenas, C. Berrospi Ramis, M. Omenetti, K. Dinkla, Y. Kim, S. Gupta, R. Teixeira de Lima, V. Weber, L. Morin, I. Meijer, V. Kuropiatnyk, and P. W. J. Staar, ``Docling: An Efficient Open-Source Toolkit for AI-driven Document Conversion,'' arXiv:2501.17887, (2025).
\bibitem{b19} Z. Zhang, H. Liu, S. Liang, Y. Zhang, Y. Xiang, J. Liu, T. Sun, M. Lin, Y. Zhang, C. Zhou, T. Gao, C. Cui, Y. Liu, D. Yu, and Y. Ma, ``PaddleOCR-VL-1.6: Expanding the Frontier of Document Parsing with Under-Optimized Region Refinement and Progressive Post-Training,'' arXiv:2606.03264, (2026).
\bibitem{b20} D. Rahnev, ``A comprehensive assessment of current methods for measuring metacognition,'' Nature Communications 16, 701 (2025).
\bibitem{b21} X. Zhong, E. ShafieiBavani, and A. Jimeno Yepes, ``Image-based table recognition: data, model, and evaluation,'' Proc. European Conference on Computer Vision (ECCV), arXiv:1911.10683, (2020).
\bibitem{b22} Alibaba Cloud, ``Qwen3.8-Flash,'' official Model Studio page, \url{https://help.aliyun.com/en/model-studio/qwen3-8-flash}, (2026).
\bibitem{b23} OpenRouter, Inc., ``OpenRouter: a unified API for language models,'' official web page, \url{https://openrouter.ai}, (2026).
\bibitem{b24} Anthropic, ``Introducing Claude Opus 5.5,'' official model page, \url{https://www.anthropic.com/claude-opus-5-5}, (2026).
\bibitem{b25} OpenAI, ``GPT-6 Luna,'' official model documentation, \url{https://developers.openai.com/api/docs/models/gpt-6-luna}, (2026).
\bibitem{b26} Anthropic, ``Introducing Claude Haiku 5.5,'' official model page, \url{https://www.anthropic.com/claude-haiku-5-5}, (2026).
\bibitem{b27} P. Cai, Z. Zou, S. Liu, Y. Wang, J. Tang, K. Yang, M. Tong, Z. He, and H. Sun, ``TeleOCR: Navigating document parsing across digital and camera-captured documents,'' arXiv:2608.12898, (2026).
\bibitem{b28} J. Zhang, B. Wang, Q. Zhang, et al., and C. He, ``TRivia: Self-supervised fine-tuning of vision-language models for table recognition,'' arXiv:2512.01248, (2025).
\bibitem{b29} S. Peng, A. Chakravarthy, S. Lee, X. Wang, R. Balasubramaniyan, and D. H. Chau, ``UniTable: Towards a unified framework for table recognition via self-supervised pretraining,'' arXiv:2403.04822, (2024).
\end{thebibliography}
\end{document}